\documentclass[conference]{IEEEtran}
\IEEEoverridecommandlockouts
\usepackage{cite}
\usepackage{amsmath,amssymb,amsfonts}
\usepackage{algorithmic}
\usepackage{graphicx}
\usepackage{textcomp}
\usepackage{xcolor}
\usepackage{hyperref}
\usepackage{xurl}
\usepackage{dblfloatfix} 

\makeatletter
\newcommand{\linebreakand}{%
  \end{@IEEEauthorhalign}
  \hfill\mbox{}\par
  \mbox{}\hfill\begin{@IEEEauthorhalign}
}
\makeatother

\def\BibTeX{{\rm B\kern-.05em{\sc i\kern-.025em b}\kern-.08em
    T\kern-.1667em\lower.7ex\hbox{E}\kern-.125emX}}
\begin{document}

\title{Does Dataset Complexity Matters for Model Explainers?\\

}

\author{\IEEEauthorblockN{1\textsuperscript{st} José Ribeiro}
\IEEEauthorblockA{\textit{Federal University of Pará - UFPA}\\
\textit{Federal Institute of Pará - IFPA}\\
 Belém, Brazil \\
jose.ribeiro@ifpa.edu.br}
\and
\IEEEauthorblockN{2\textsuperscript{nd} Raíssa Silva}
\IEEEauthorblockA{\textit{IRMB, Montpellier University} \\
La Ligue Contre le Cancer \\
Montpellier, France\\
r.lorenna@gmail.com}
\and
\IEEEauthorblockN{3\textsuperscript{nd} Lucas Cardoso}
\IEEEauthorblockA{\textit{ICEN, Federal University of Pará}
\\Belém, Brazil\\
lucas.cardoso@icen.ufpa.br}
\linebreakand
\IEEEauthorblockN{4\textsuperscript{rd} Ronnie Alves}
\IEEEauthorblockA{\textit{Federal University of Pará - UFPA}\\
\textit{Vale Institute of Technology - ITV DS}\\
Belém, Brazil \\
ronnie.alves@itv.org}
}

\maketitle

\begin{abstract}
Strategies based on Explainable Artificial Intelligence - XAI have emerged in computing to promote a better understanding of predictions made by black box models. Most XAI measures used today explain these types of models, generating attribute rankings aimed at explaining the model, that is, the analysis of Attribute Importance of Model. There is no consensus on which XAI measure generates an overall explainability rank. For this reason, several proposals for tools have emerged (Ciu, Dalex, Eli5, Lofo, Shap and Skater). An experimental benchmark of explainable AI techniques capable of producing global explainability ranks based on tabular data related to different problems and ensemble models are presented herein. Seeking to answer questions such as ``Are the explanations generated by the different measures the same, similar or different?" and ``How does data complexity play along model explainability?" The results from the construction of 82 computational models and 592 ranks shed some light on the other side of the problem of explainability: dataset complexity!

\end{abstract}


\section{Introduction}
Recently, technology has increasingly evolved and allowed intelligent algorithms to be present in our daily lives through solutions to the most diversified types of problems, thus prompting a need for machine learning models to solve increasingly complex problems which require characteristics that justify decision making, in addition to high problem-solving performance\cite{shalev2014understanding}\cite{ghahramani2015probabilistic}.

Computational models based on bagging and boosting algorithms, which provide for high performance and high generalization capacity, are commonly used in computation to solve regression and classification problems based on tabular data. However, these models are not considered transparent algorithms\footnote{Transparent Algorithms: Algorithms that generate explanations for how a given output was produced. Such examples include Decision Tree, Logistic Regression and K-nearest Neighbors.}, being considered black box algorithms\footnote{Black Box Algorithms: Machine learning algorithms that have classification or regression decisions that are hidden from the user.} and, therefore, less used in problems related to sensitive contexts, such as health and safety \cite{arrieta_explainable_2019_20}\cite{explainable_bigdata_health}\cite{explainable_bigdata_safety}.

The limited understanding of black box models requires a search for measures or tools that can provide information about local explanations --- aiming to predict around an instance through various methods to obtain a local attribute importance ranking \cite{xai_local_global_2020} ---  and global explanations --- when it is possible to understand the logic of all instances of the model by generating a global attribute importance rank \cite{xai_local_global_2020}\cite{guidotti2018survey} --- as a means of making decisions interpretable and, thus, more reliable \cite{darpa_2019}.

Efforts have been made in the Explainable Artificial Intelligence - XAI area regarding the development of different measures to explain black box models, even after their training and testing process, called post-hoc analysis \cite{arrieta_explainable_2019_20}. Thus, measures such as Ciu \cite{ciu_ref}, Dalex\cite{dalex_book}, Eli5\cite{eli5_ref}, Lofo\cite{lofo_ref}, Shap\cite{tree_shap_ref}, and Skater\cite{skater_ref} emerged to promote the creation of model-agnostic\footnote{Model-Agnostic: means it does not depend on the type of model to be explained.} explanations. Each of these tools is capable of generating explanations using different techniques, but a fact they have in common is that they all generate global attribute importance rankings related to the explanation of a model.

The proposal to analyze the ranking of global importance --- and not the local --- allows general analysis of how a given model treats and explains the problem to be solved, and for this reason, it was chosen for this study.

This research raises discussions for the current moment in the XAI area through two main questions, namely: --- Considering the current measures geared at explaining black box machine learning models, can it be inferred that they generate global rankings of same, similar, or different explainabilities? --- Following the same idea as in the previous question, are the generations of equal, similar, or different explainabilities related to specific properties of a dataset?

Seeking to answer the two hypotheses above, this research emerges as a comparative analysis of different XAI measures, capable of producing global explainability ranks based on tabular data related to different problems and ensemble models.

The main contributions of this research to the machine learning area focused on XAI are as follows:

\begin{itemize}
\item Development of a benchmark that measures the correlations existing between the explanations generated by the main current XAI measures by comparing them with each other through different perspectives (properties of the datasets used);

\item Generation of comparative results between explainability ranks that lead researchers in the XAI area to identify which explainability measures to use against different characteristics of a dataset.
\end{itemize}

\section{Background}
\subsection{Explainable Artificial Intelligence}

Recently, there is a growing need to explain models of black box machine learning in an agnostic way. This includes not only more robust computational models like Deep Neural Networks but also simpler models like ensembles \cite{explicabilidade_rede_neural}\cite{kindermans_learning_2017_2018}\cite{review_xai_2021}\cite{ethical_ml_git}.

In this way, the herein research identifies in the current literature the necessity of specific studies on model-agnostic XAI measures for tabular data, thus enabling explanations of computational models with wide applicability, such as ensemble trees \cite{review_xai_2021}\cite{arrieta_explainable_2019_20}.

A bibliographic and practical survey (development) on the main existing XAI measures was performed through this research, specifically aimed at generating model-agnostic global explainability ranks that support tabular data. As a result, a total of six tools were selected, with updated and compatible libraries with current Python development platforms and Scikit-Learning algorithms \cite{scikit-learn}. These tools are: CIU \cite{ciu_ref}, Dalex \cite{dalex_r_ref}, Eli5 \cite{eli5_ref}, Lofo \cite{lofo_ref}, SHAP \cite{tree_shap_ref} and Skater \cite{skater_ref}.

Note that all the six measures referenced herein generate explanatory ranks based on the same machine learning models previously trained, manipulate their inputs or/and produce new intermediate models (copies). Therefore, a comparison of the generated explanatory ranks is fair and feasible.

Due to the incompatibility between the current XAI measurement libraries and the versions of the \textit{scikit-lean} \cite{scikit-learn} computational model libraries and dependencies, only the Random Forest and Gradient Boosting algorithms were used in this study.

During the initial analyzes and executions, this research found XAI measures with incompatibilities at the level of libraries and dependencies, which made it impossible to create uniform tests with the other measures. For this reason, some XAI means were not used for this research (for example: Alibe-ALE \cite{alibi_ale_ref}, Lime \cite{lime_ref}, Ethical ExplainableAI \cite{ethical_xai_site}, IBM Explainable AI 360 \cite{ibm_xai360}, and Interpreter ML \cite{interpretML_arxiv}).

Contextual Importance and Utility - CIU is a XAI measure based on Decision Theory \cite{decisions_1993} that focuses on serving as a unified measure of model-agnostic explainability based on the implementation of two different indexes, namely: Contextual Importance - CI and Contextual Utility - CU, which generally create equal attribute ranks, but with different values for each index, thus generating global explanations\cite{ciu_ref}. Only the CI is used in this research, for obeying the same CU rank sequence.

Dalex is a set of XAI tools based on the LOCO (leave-one covariate out) approach and it can generate explainabilities from this approach. In general, the measure receives the model and the data to be explained; it calculates the performance of the model, performs new training processes with new generated data sets, and it does the inversion of each attribute of the data in a unitary and iterative way, thereby measuring which attributes are most important to the model based on the performance  \cite{dalex_python_ref} \cite{dalex_book}.

Leave One Feature Out - Lofo is a XAI measure that has a similar proposal than Dalex, but with a main difference regarding the inversion of the attribute iterativity, because in the Lofo measure the iterative step is the removal of the attribute to find its global importance to the model based on the performance \cite{lofo_ref}.

Explain Like I’m Five - Eli5 is a tool that helps explore machine learning classifiers and explains their predictions by assigning weights to decisions, as well as exporting decision trees and presenting the importance of the attributes of the model submitted to the tool \cite{eli5_ref}.

SHapley Additive exPlanations - SHAP is proposed as a unified measure of attribute importance that explains the prediction of an instance $X$ from the contribution of an attribute. The contribution of this attribute is calculated from the game theory of Shapley Value \cite{roth1988shapley}. In this tool, it is possible to calculate the average of the Shapley Value values by attributes to obtain the global importance \cite{tree_shap_ref}\cite{kernel_shap_ref}.

Skater is a set of tools capable of generating rankings of the importance of model attributes, based on Information Theory \cite{info_theory_1994}, through measurements of entropy in changing predictions, through a perturbation of a certain attribute. The central idea is that the more a model is dependable upon an attribute, the greater the change in predictions\cite{skater_ref}.

A general comparison of the main characteristics of the measures that meet the requirements of this research is presented in table \ref{tab_resumo_xai}.

\begin{table*}[h]
\centering
\caption{Main XAI measures surveyed}
\resizebox{.8\textwidth}{!}{%
\begin{tabular}{|c|c|c|c|c|c|}
\hline
\textbf{\begin{tabular}[c]{@{}c@{}}XAI \\ meansure\end{tabular}} & \textbf{Autor} & \textbf{Algorithm basis} & \textbf{\begin{tabular}[c]{@{}c@{}}Global \\ explanation\\ (by rank)\end{tabular}} & \textbf{\begin{tabular}[c]{@{}c@{}}Local \\ explanation\end{tabular}} & \textbf{\begin{tabular}[c]{@{}c@{}}API\\ compatible\end{tabular}} \\ \hline
CIU & \cite{ciu_ref} & Decision Theory & Yes & No & Yes \\ \hline
Dalex & \cite{dalex_r_ref} & Leave-one covariate out & Yes & Yes & Yes \\ \hline
Eli5 & \cite{eli5_ref} & Assigning weights to decisions & Yes & Yes & Yes \\ \hline
Lofo & \cite{lofo_ref} & Leave One Feature Out & Yes & No & Yes \\ \hline
SHAP/Tree & \cite{tree_shap_ref} & Game Theory & Yes & Yes & Yes \\ \hline
Skater & \cite{skater_ref} & Information Theory & Yes & Yes & Yes \\ \hline
\end{tabular}%
}
\label{tab_resumo_xai}
\end{table*}

\begin{figure*}[b]
\begin{center}
\includegraphics[scale=0.35]{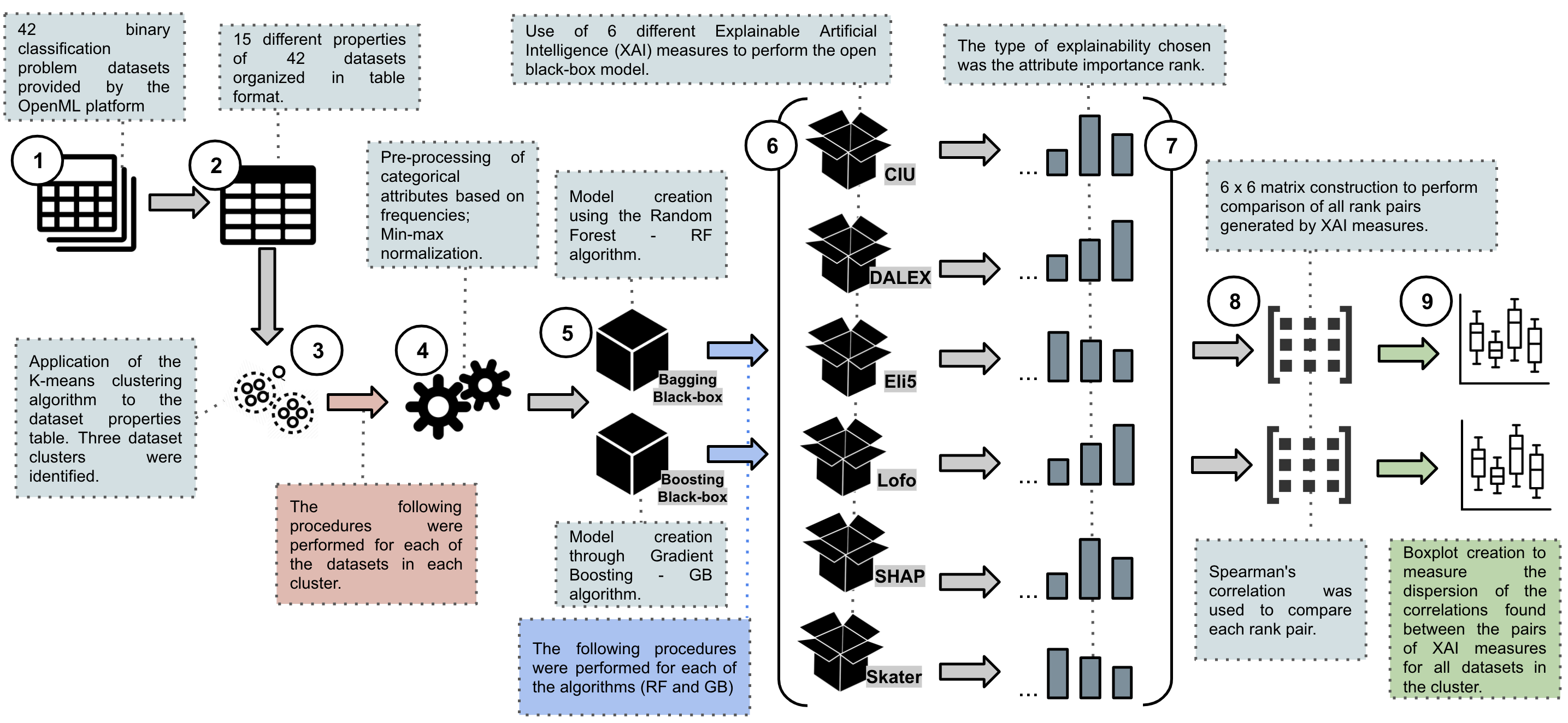}
\caption{Visual scheme of all steps and processes performed by the proposed benchmark.}
\label{fig_metodologia}
\end{center}
\end{figure*}

All the tools described above are capable of generating explainabilities that go beyond the generation of attribute importance ranks; therefore, the focus of this article is to compare this most basic unit of model explainability, i.e., the global importance rank.

With the explanatory ranks generated by the different measures, it is fair to compare existing correlations between these ranks, even if each measure generates them based on different algorithms, since the central objective of each measure is to explain a black box model.

\subsection{Dataset Properties}

After being processed through attribute engineering, a dataset is expected to exhibit characteristics that involve the nature of the problem in a contextual (problem properties) and technical (dataset properties) manner and which are generalized by machine learning algorithms to carry out a prediction or a classification process \cite{darpa_2019}. 

Although the contextual side of the problem significantly helps in interpreting the explainabilities of the models, the herein  researchers opted not to work with these characteristics of the analyzed data, since datasets utilized refer to different problems with contexts that often go beyond computing knowledge.

On the other hand, there are properties of datasets which are feasible to be analyzed, since they are common --- even though these datasets represent different problems. Properties such as dimensionality, number of numerical attributes, number of binary attributes, the balance between classes, and entropy are examples of properties that every tabular dataset has and which directly influence the generalizability of the proposed model \cite{dataset_characteristics_effects}.

By the time this article was published, research that confronted the inherent characteristics of different datasets (complexities) and their explanations through machine learning models was not identified in the literature. In this sense, this research seeks to fill in this gap in the framework of XAI measure studies.

\section{Materials and Methods}

This research developed a benchmark, Figure \ref{fig_metodologia}, which uses tabular data already known and endorsed by the machine learning community (Figure \ref{fig_metodologia} - 1), extracted properties from these datasets (Figure \ref{fig_metodologia} - 2), clustered these datasets according to their properties (Figure \ref{fig_metodologia} - 3), performed the construction of computational models (Figure \ref{fig_metodologia} - 4 and 5), generated explainability ranks for all models (Figure \ref{fig_metodologia} - 6 and 7), calculated the correlations obtained from the all rank pairs and analyzed the different results (Figure \ref{fig_metodologia} - 8 and 9). All these procedures are explained in detail in the following topics.   

\subsection{Datasets and Preprocess}

A total of 41 datasets of binary problems were used, with no data loss, and selected from the OpenML \cite{openml} platform. The most used datasets were selected, so as to make the results of this research even better for use by the machine learning community.

The datasets used were as follows: \textit{australian, phishing websites, spec, satellite, analcatdata lawsuit, banknote authentication, blood transfusion service center, churn, climate model simulation crashes, credit-g, delta ailerons, diabetes, eeg-eye-state, haberman, heart-statlog, ilpd, ionosphere, jEdit-4.0-4.2, kc1, kc2, kc3, kr-vs-kp, mc1, monks-problems-1, monks-problems-2, monks-problems-3, mozilla4, mw1, ozone-level-8hr, pc1, pc2, pc3, pc4, phoneme, prnn crabs, qsar-biodeg, sonar, spambase, steel-plates-fault, tic-tac-toe} and \textit{wdbc}.

All datasets went through the following pre-processing steps, as required: converting categorical attributes to frequency-based ordinals, converting boolean attributes to integer (0 and 1), and min-max normalization to values between 0 and 1 \cite{sklearn_minmaxscaler}.

\subsection{Clustering and Multiple Correspondence Analysis}

Analyzing different datasets based on their properties allows for comparing similarities and differences between them, even if these datasets are from different contexts. For example, we can group a set of datasets based on their class entropy values and then identify which datasets have more and less information, thus enabling a dataset complexity impact relationship for future model explanations.

Based on the proposal laid out above, this research used 15 different properties (provided by OpenML) extracted from each of the 41 selected datasets and then used the clustering algorithm k-means \cite{kmeans} to identify groups of datasets based on their similarities. 

\begin{figure}[h]
\begin{center}
\includegraphics[scale=0.6]{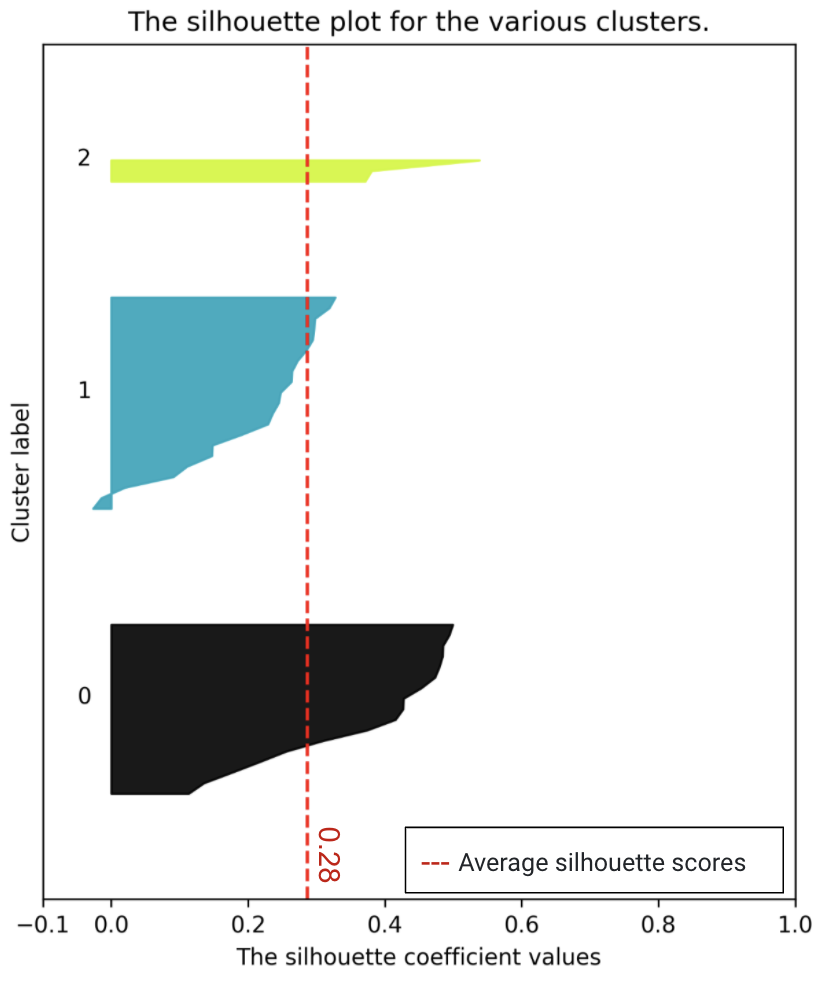}
\caption{Silhouette coefficients for clustering, using the Kmeans algorithm, for a $=3$. Distance means ($x$ axis) and label of clusters $0,1$ and $2$ (axis $y$).}
\label{fig_silhuette}
\end{center}
\end{figure}

The interpretation and validation algorithm between data clusters, called Silhouettes \cite{silhouettes}, was used, the K value (clusters) of which ranging between 2 and 20, and K = 3 with Average Silhouette Scores = 0.28 being identified, Figure \ref{fig_silhuette}.

In figure \ref{fig_silhuette}, it can be seen that for a k=3 the distances between each cluster (0, 1 and 2) are above the average (red line), so this is an adequate value of k in relation to the study presented hereby.

The following 15 properties being used included: \textit{Number of Features, Number of Instances, Dimensionality, Percentage of Binary Features, Standard Deviation Nominal of Attribute Distinct Values, Mean Nominal Attribute Distinct Values, Class Entropy, Autocorrelation, Number of Numeric Features, Number of Symbolic Features, Number of Binary Features, Percentage of Symbolic Features, Percentage of Numeric Features, Majority Class Percentage}, and \textit{Minority Class Percentage}.

Further information about the values of each property for each dataset can be found in: \url{https://github.com/josesousaribeiro/XAI-Benchmark/blob/main/Openml/df_dataset_properties.csv}.

In order to identify the different correlations between each dataset cluster identified with the 15 properties listed above, this research used the Multiple Correspondence Analysis - MCA \cite{ref_mca} in the properties dataset, where the rows of this table are the observations or individuals $(n)$ concerned --- here are the datasets --- and the columns are the different categories of nominal variables $(p)$ --- here are the properties of each dataset.

This research performed the process of binarization in the same table in which clustering was applied, replaced by $h$ (equal to or above the mean of attribute values) and $s$ (below the mean of attribute values). For more information about the new properties dataset, refer to: \url{https://github.com/josesousaribeiro/XAI-Benchmark/blob/main/Openml/df_properties_binarized.csv}.

This analysis allows for identifying different relationships between dataset (or even dataset group) and properties (or even property value ranges), as advocated by the literature \cite{ref_mca}.

\subsection{Construction of Models}

The algorithms used in building each dataset model were Random Forest - RF and Gradient Boosting - GB, both being tree-based, low-explainability ensembles \cite{arrieta_explainable_2019_20}.

This research tested newer ensemble algorithms such as Light Gradient Boost \cite{lightgbm}, CatBoost \cite{catboost}, and Extreme Gradient Boosting \cite{xgboost}; however, not all XAI metrics that were used supported the output encodings of these new models. Thus, an option was made to adhere to the Gradient Boosting and Random Forest algorithms, only.

Notably, the Tuning step was performed through Grid Search \cite{gridsearchcv} with cross-validation \cite{crossvalidation_2020} of $fold = 3$, thus that the algorithms had a better calibration on the analyzed data, and a total of 192 candidates through the parameters: \textit{max depth: (1, 10), bootstrap: (True, False), n estimators: (100, 200), min samples leaf: (1, 10), min samples split:(2,10), max features: (sqrt, log2)}, and \textit{loss: (deviance, exponential)}  (the latter only valid for the GB algorithm).

Finally, each model was trained and tested respectively based on a $70\% - 30\%$ ration for each dataset. Then, each model was measured as to their performance and stability regarding accuracy and precision, recall and Friedman test \cite{noauthor_friedmantest_2020}, so as to characterize each model that was created.

\subsection{XAI Ranks and Correlations}

For each of the 41 datasets (divided into 3 different clusters), 2 models were created, one based on the Random Forest algorithm and the other on Gradient Boosting, thus generating a total of 82 models. Global explainability ranks were produced through the 6 different XAI measures for each model, resulting in a total of 492 ranks.

\begin{figure}[h]
\begin{center}
\includegraphics[scale=0.6]{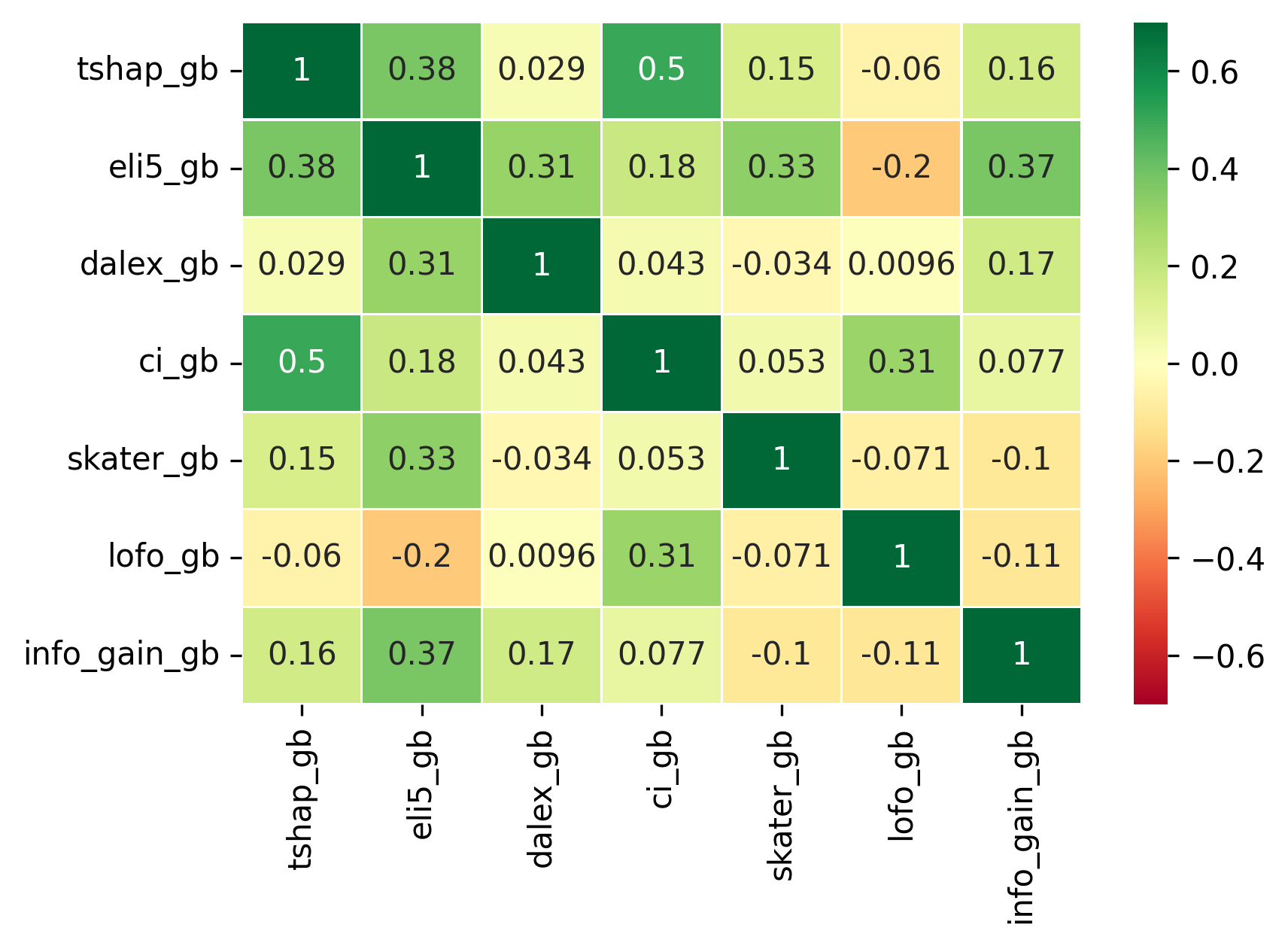}
\caption{Example Spearman correlation matrix (represented in value and color in each cell) based on explicability ranks generated from Gradient Boosting model to  the dataset $wdbc$. Note that, for this dataset, chosen at random, the correlations between the rank pairs are low or non-existent (close to zero).} 
\label{fig_correlacoes}
\end{center}
\end{figure}

The Spearman Rank Correlation \cite{spearman_ref} was used to calculate the correlations between the ranks that were created. The reason for using this algorithm in particular, is that it measures the correlation between rank pairs considering the idea of ranks (positions) in which different values (in this case, dataset attributes) may appear. In this step, two comparison matrices of rank correlation pairs are generated for each dataset (one matrix for each algorithm). Figure \ref{fig_correlacoes} shows an example of this matrix created from the Gradient Boosting - GB model.

Correlations such as those shown in Figure \ref{fig_correlacoes} are calculated for each dataset and show how the correlations between different pairs of XAI measures are presented for each problem and for each computational model being developed.

The results shown in Figure \ref{fig_correlacoes} refer to only one dataset and, therefore, are intermediate results, since they will be summarized in the results of the other datasets.

\subsection{Benchmark of XAI measures}

All the methodology described so far was consolidated and implemented in a single benchmark. More details about the implementation and execution of the procedures presented hereby are available in the repository: --- \url{https://github.com/josesousaribeiro/XAI-Benchmark}.

\section{Results and Discussion}

The results obtained were computed in groups (2 different computational models) and clusters (3 different dataset clusters).

The principle for understanding the results of this article is to acknowledge the different dataset cluster profiles identified in the clustering process and MCA. 

In this sense, according to clustering, based on the 15 different properties of the 41 datasets that were analyzed, it is possible to assess the existence of at least 3 different dataset clusters. In quantitative terms, each cluster has 21 (cluster 0), 17 (cluster 1), and 3 (cluster 2) datasets.

To identify each different dataset profile, an inspection of the value ranges of each cluster for each property was carried out. This analysis verified the possible existence of specific property values for datasets of specific clusters.

At this stage, it was identified that the quantity of datasets in cluster 2 had a small $n$ and, therefore, it was disregarded from these results (the datasets for this cluster are: \textit{kr-vs-kp, Phishing Web sites,} and \textit{SPECT}). Keeping only clusters 0 and 1.

Intending to consolidate the profile analysis of the clusters, Multiple Correspondence Analysis - MCA was applied, and the relation between the datasets and the value ranges of their properties was verified, thus providing for a better understanding of the complexity of the datasets existing in both clusters --- this practice is already known in the literature \cite{ref_mca}.

\begin{figure*}[h]
\begin{center}
\includegraphics[scale=0.8]{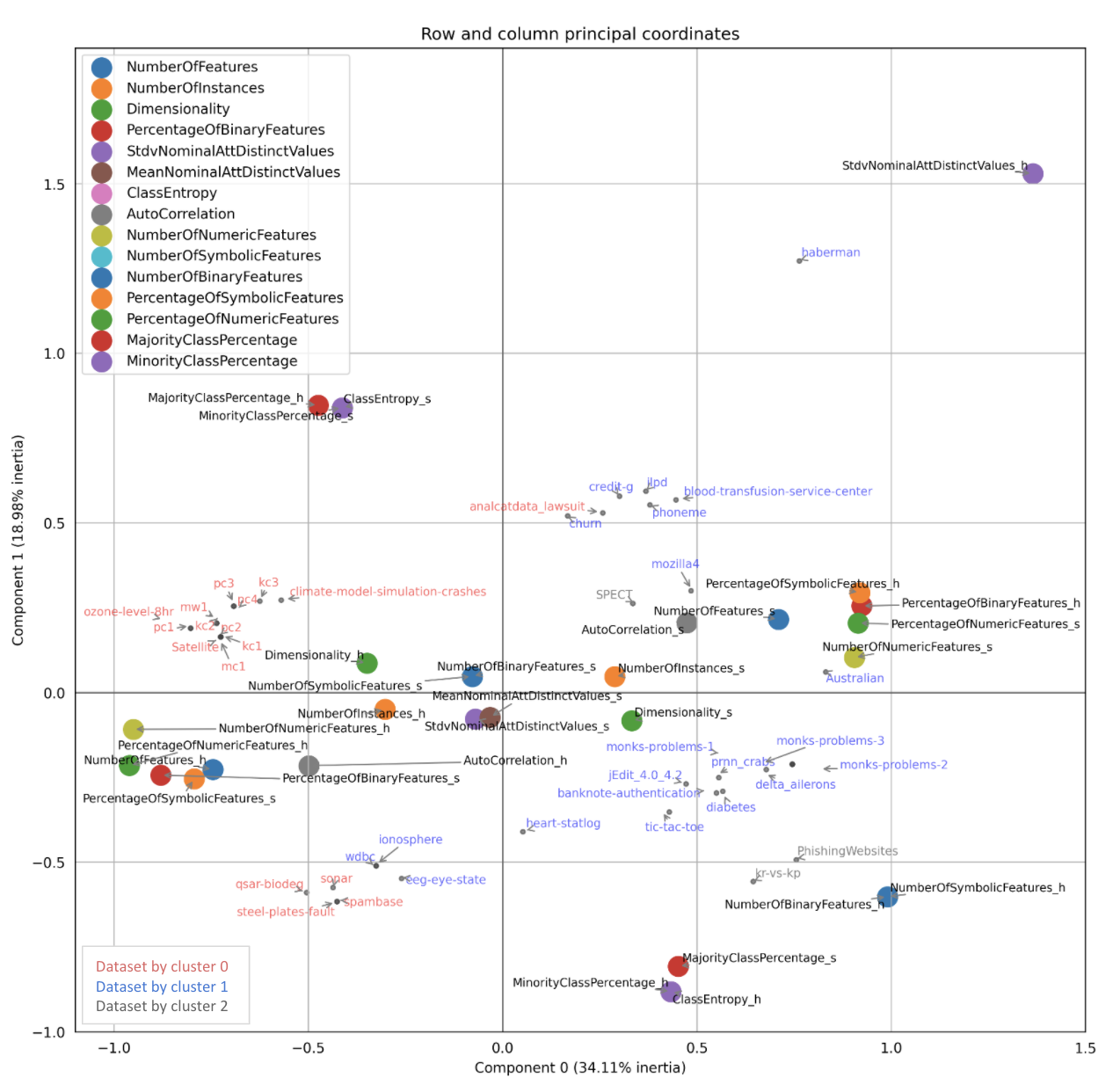}
\caption{Multiple Correspondence Analysis - MCA with rows (datasets) and columns (properties) depending on the first and second PCA components.}
\label{fig_mca}
\end{center}
\end{figure*}

By inspecting the MCA shown in Figure \ref{fig_mca}, the relation (shorter distances) between high quantity of datasets in cluster 0 and above-average values (symbol $h$) for properties (bigger circles) such as \textit{MajorityClassPercentage, Dimensionality, NumberOfInstances, NumberOfFeatures, PercentageOfNumericFeatures}, and \textit{AutoCorrelation} was identified. Also, a relation between a high quantity of datasets of cluster 0 and below-average values (symbol $s$) for properties such as \textit{ClassEntropy, PercentageOfSymbolicFeatures}, and \textit{PercentageOfBinaryFeatures} was observed.

Also with regard to the MCA shown in Figure \ref{fig_mca}, by inspection, assessing the existence of a relation (shorter distances) between a high quantity of datasets in cluster 1 and above-average values (symbol $h$) is possible for properties (bigger circles) such as \textit{ClassEntropy, PercentageOfSymbolocFeatures, PercentageOfBinaryFeatures, NumberOfBinaryFeatures}, and \textit{MinorityClassPercentage}. It is also possible to identify a relation between a high quantity of datasets in cluster 1 and below-average values (symbol $s$) for properties sush as \textit{AutoCorrelation,NumberOfFeatures, PercentageOfNumericFeatures, NumberOfNumericFeatures, NumberOfInstances, Dimensionality}, and \textit{MajorityClassPercentage}.

It is worth noting that some properties were not mentioned above because they appear at very close distances for the two dataset clusters.

However, based on the inspection of MCA results, it can be concluded that cluster 0 is formed by a significant amount of datasets with greater complexities, whereas cluster 1 is formed by a significant amount of datasets with minor complexities.

The entire cluster profile identification process can be found in: \url{https://github.com/josesousaribeiro/XAI-Benchmark/blob/main/Cluster_Profile.md}.

Based on the profile characterization of the two main clusters that were identified, this research executed the Benchmark for these two sets of datasets separately.

\begin{figure*}
\includegraphics[width=\textwidth]{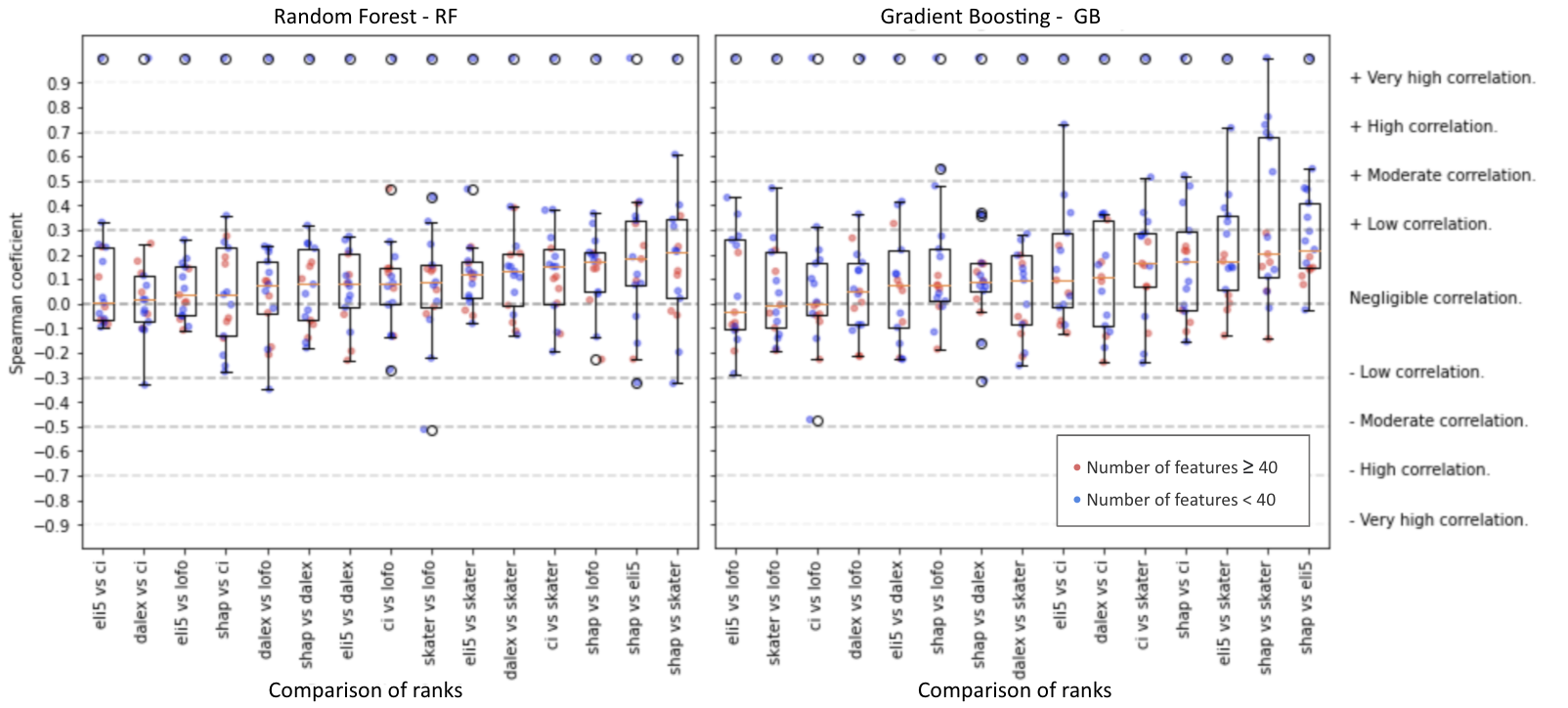}
\caption{Boxplot graphs that summarize all rank pair comparisons ($x$ axis) and their respective Spearman correlations ($y$ axis) calculated for the datasets of cluster 0. Results of the model based on Random Forest (left) and Gradient Boosting (right). The dashed blue lines show the different levels of correlations. The points identify the positions of the correlation values of each comparison, and the color of the points refer to the attribute quantities of each dataset. Note the low variance of most boxplots along with levels of negligent or low correlations found.}
\label{fig_boxplot_cluster_0}
\end{figure*}

The benchmark execution for the datasets belonging to cluster 0, Figure \ref{fig_boxplot_cluster_0}, exhibits low or negligible correlations found in most tests, thus showing that the six XAI measures used hereby generate different explainability ranks for most of the datasets in this group, regardless of the algorithm (RF or GB) the computational model is based upon.

\begin{figure*}
\includegraphics[width=\textwidth]{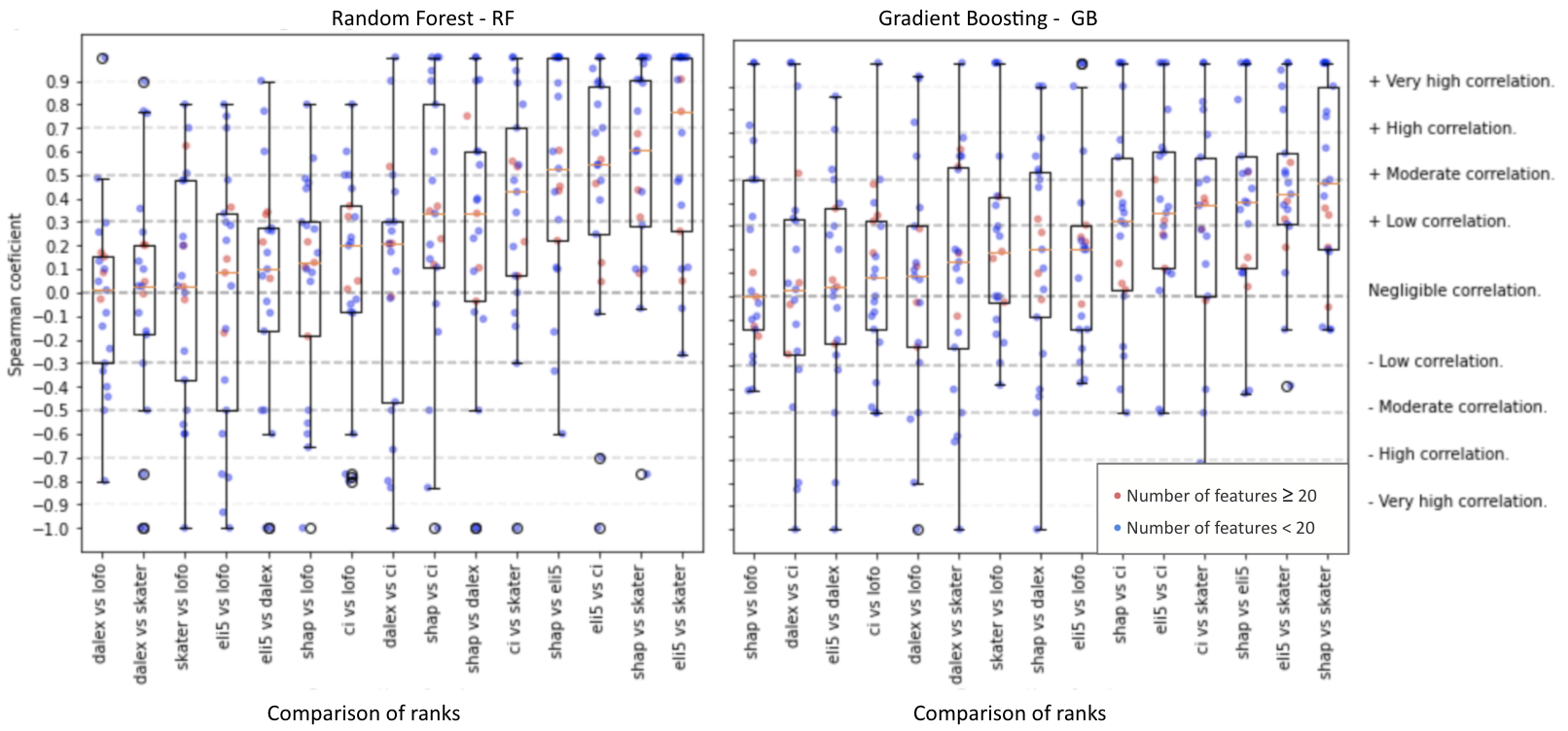}
\caption{Boxplot graphs that summarize all rank pair comparisons ($x$ axis) and their respective Spearman correlations ($y$ axis) calculated for the datasets of cluster 1. Results of the model based on Random Forest (left) and Gradient Boosting (right). The dashed blue lines show the different levels of correlations. The points identify the positions of the correlation values of each comparison, and the color of the points refer to the attribute quantities of each dataset. Note the high variances of most boxplots along with the high levels of correlations found.} 
\label{fig_boxplot_cluster_1}
\end{figure*}

The results presented by cluster 0, Figure \ref{fig_boxplot_cluster_0}, show that datasets with greater complexity generate more complex models, and in turn, different explanations. This could justify the low correlations found in the results.

The benchmark execution for the datasets of cluster 1, Figure \ref{fig_boxplot_cluster_1}, shows different correlations from those shown in Figure \ref{fig_boxplot_cluster_0}, since there is high variance in boxplots as well as high positive and negative correlations between rank pairs in cluster 1.

The results presented in Figure \ref{fig_boxplot_cluster_0} and \ref{fig_boxplot_cluster_1} provide evidence on how to answer the first hypothesis launched herein: \textit{``--- Considering the current measures aimed at explaining black box machine learning models, can it be inferred that they generate global rankings of same, similar, or different explainabilities?''}. It depends, since the different correlations found demonstrate that the properties of the datasets directly influence the explainabilities of the models created from them. Thus, it is possible to make the tools generate explainabilities with higher correlations with one another (more similar explainabilities) and lower correlations with one another (less similar explainabilities).

Noticeably, despite the similar results of the models based on the RF and GB algorithms, there are important differences in the correlations found, as shown in Figure \ref{fig_boxplot_cluster_0}. For example, the larger variance of some comparisons in the results of the GB algorithm (\textit{shap vs lofo, eli5 vs ci, eli5 vs skater, ci vs skater} and \textit{shap vs ci}). This proves that the complexity of the algorithm used in the model (in this case, bagging and boosting tree-based) does have an influence on explainability generation. 

Importantly, the outliers existing in the boxplots of Figures \ref{fig_boxplot_cluster_0} and \ref{fig_boxplot_cluster_1}, show the possibility of other dataset clusters existing amongst those being analyzed. The colors, used to distinguish the amount of attributes from different datasets in each boxplot show that the results obtained in the correlations are not a function of the number of attributes that a dataset has.

The comparison of the results presented in Figures \ref{fig_boxplot_cluster_0} and \ref{fig_boxplot_cluster_1} demonstrate the answer to the second hypothesis of this research, namely: \textit{``--- According to the same idea as in the previous question, are the generations of equal, similar, or different explainabilities related to specific properties of a dataset?''}. Yes, since the results of the rank correlations for the two dataset clusters presented in the benchmark output were different, thus  showing that the complexity properties of the dataset interfere with explainability ranks.

Based on the experiments performed with the 6 different XAI measures and the resulted obtained therefrom, the following logical reasoning could be constructed: 1 - If a ensemble model (algorithm and dataset) solves a classification problem for a low-complexity dataset, there must be few ranks (or even only one rank) of explainabilities referring to this model, thus allowing to infer that the different XAI measures have higher correlations with each other (as seen in Figure \ref{fig_boxplot_cluster_1}); 2 - If an ensemble model (algorithm and dataset) solves a classification problem for a high-complexity dataset, there must be many explainability ranks for this model, leading the ranks of the different XAI measures to show lower correlations with each other (as seen in Figure \ref{fig_boxplot_cluster_0}).

Based on the logical reasoning 1 and 2 above, there was identified existence of evidence about Ensemble-Tree Models with Complex Explainabilities - EMCX (created from the data of cluster 0) and Ensemble-Tree Models with Simple Explainabilities - EMSX (created from the data of cluster 1), both terms proposed hereby. Furthermore, it was identified that the divergence between the ranks is inherent, primarily from the complexity of the dataset and, consequently, from the generated model --- thus answering the article's title question.

The datasets depicted below, which served as the basis for the creation of EMSX and EMCX, are intended to comply with one of the contributions hereto:

\begin{itemize}

\item EMSX Datasets: \textit{Ozone level 8hr, Sonar, Spambase, Qsar biodeg, Kc3, Mc1, Pc3, Mw1, Pc4, Satellite, Pc2, Steel plates fault, Kc2, Pc1, Kc1, Climate model simulation crashes, and Analcatdata lawsuit};

\item EMCX Datasets: \textit{Ionosphere, Wdbc, Credit-g, Churn, Australian, Eeg eye state, Heart statlog, Ilpd, Tic tac toe, JEdit 4.0 4.2, Diabetes, Prnn crabs, Monks problems 1, Monks-problems 3, Monks problems 2, Delta ailerons, Mozilla4, Phoneme, Blood transfusion service center, Banknote authentication, and Haberman};

\end{itemize}

It is clear that evidence of the existence of EMSX and EMCX was observed from experiments with ensemble algorithms. For these terms to become generally applicable (other algorithms), experiments with algorithms of other natures are necessary.

\section{Final Considerations}

Given all the analyzes that were carried out, this research achieves its objective of being able to observe the impacts of dataset complexity in explaining black box machine learning models, through the different properties of the data under analysis. That allowed for finding that the properties of a dataset --- facing the creation of models based on Random Forest and Gradiant Boosting algorithms --- show evidence that EMSX and EMCX do exist. However, many other properties of the data and of other algorithms still need to be further explored.



\bibliographystyle{IEEEtran}
\bibliography{mylib}

\end{document}